\documentclass{article}

\usepackage[preprint]{neurips_2025}

\usepackage[utf8]{inputenc} 
\usepackage[T1]{fontenc}    
\usepackage{hyperref}       
\usepackage{url}            
\usepackage{booktabs}       
\usepackage{amsfonts}       
\usepackage{nicefrac}       
\usepackage{microtype}      
\usepackage{xcolor}         

\usepackage{amsmath,amsfonts,bm,amsthm}
\usepackage{algorithm}
\usepackage{algorithmic}
\usepackage{amsmath}
\usepackage{graphicx}
\usepackage{subcaption}
\DeclareUnicodeCharacter{3001}{\'{e}}
\usepackage{natbib}
\setcitestyle{numbers,square}

\title{NSFW-Classifier Guided Prompt Sanitization for Safe Text-to-Image Generation}

\author{%
  Yu Xie\thanks{Equal contribution.} \\
  Purple Mountain Laboratories\\
  \texttt{xieyu01@pmlabs.com.cn} \\
  \And
  Chengjie Zeng\footnotemark[1] \\
  Fudan University \\
  \texttt{cjzeng24@m.fudan.edu.cn} \\
  \AND
  Lingyun Zhang \\
  Fudan University  \\
  \texttt{lyzhang22@m.fudan.edu.cn} \\
  \And
  Yanwei Fu \\
  Fudan University \\
  \texttt{yanweifu@fudan.edu.cn} \\
}

\begin{document}

\maketitle

\begin{abstract}

  The rapid advancement of text-to-image (T2I) models, such as Stable Diffusion, has enhanced their capability to synthesize images from textual prompts. However, this progress also raises significant risks of misuse, including the generation of harmful content (e.g., pornography, violence, discrimination), which contradicts the ethical goals of T2I technology and hinders its sustainable development. Inspired by "jailbreak" attacks in large language models, which bypass restrictions through subtle prompt modifications, this paper proposes NSFW-Classifier Guided Prompt Sanitization (PromptSan), a novel approach to detoxify harmful prompts without altering model architecture or degrading generation capability. PromptSan includes two variants: PromptSan-Modify, which iteratively identifies and replaces harmful tokens in input prompts using text NSFW classifiers during inference, and PromptSan-Suffix, which trains an optimized suffix token sequence to neutralize harmful intent while passing both text and image NSFW classifier checks. Extensive experiments demonstrate that PromptSan achieves state-of-the-art performance in reducing harmful content generation across multiple metrics, effectively balancing safety and usability.
  
\end{abstract}

\section{Introduction}
\label{sec:intro}


With the continuous advancement of text-to-image models, such as the Stable Diffusion model~\cite{HoJA20, Rombach_2022_CVPR, RuizLJPRA23, saharia2022photorealistic, Podell2024SDXL}, their ability to synthesize images based on textual descriptions has gradually improved. On one hand, this promotes the practical application of text-to-image models, but on the other hand, it also increases the risk of their misuse in generating harmful images. Text-to-image models are being abused to produce pornographic, horrific, violent, discriminatory, and other disturbing images, which go against their original design intent and hinder their further development. As a result, researchers have begun focusing on how to ensure the content safety of text-to-image models.

The primary reason text-to-image models can generate harmful content lies in their training data. Current models, such as those trained on the LAION dataset~\cite{SchuhmannLAION400M21, SchuhmannLAION5B22}, learn from vast amounts of data that inevitably contain harmful or biased material. As a result, these models inherit the ability to synthesize unsafe content. Research focuses on three main approaches to mitigate this issue:(1) Data Cleansing: Filtering out harmful images from training datasets before model training. For example, Stable Diffusion 2.0 employs this strategy to reduce exposure to unsafe content; (2)Safety Fine-Tuning: Adjusting pre-trained models to suppress harmful responses. Techniques include concept ablation\cite{kumari2023ablating} and DPO-based algorithms~\cite{HoffmannDPO23, StiennonNIPS20}; (3)Generation-Phase Filtering \& Correction: Blocking unsafe prompts during inference using NSFW classifiers (e.g., Stable Diffusion and Midjourney) or dynamically guiding outputs via methods like SLD~\cite{SchramowskiSafeWorkshop22, schramowski2023safe} and ESD-u~\cite{ZhangESDu23}, which modify noise prediction in training or inference. However, current solutions have trade-offs: data cleansing and safety fine-tuning~\cite{kumari2023ablating} may weaken generative capabilities, while input filtering~\cite{RameshGLIDE22} and service denial degrade user experience. 

We argue that text-to-image models require large amounts of data for training, and even if the training data contains harmful content, it can improve the model's image generation quality. The key to mitigating the misuse of such models for generating harmful images lies in reducing the input of harmful prompts. Inspired by jailbreak attacks~\cite{WeiZhou23, schwinn2025softpromptthreatsattacking, Yi2024survey, PerezRibeiro2022} on large language models (LLMs), which only require modifying a few input prompt tokens or adding a small number of tokens to alter the model's output response, without affecting the model's generative capabilities, this paper proposes the NSFW-Classifier Guided Prompt Sanitization method (PromptSan). This approach purifies harmful content in input prompts by modifying or appending additional tokens.

Specifically, we introduce two variants of the NSFW-Classifier Guided Prompt Sanitization method: PromptSan-Modify and PromptSan-Suffix. \textbf{PromptSan-Modify} operates during the model's inference phase. It uses a text NSFW classifier to locate harmful tokens in the input prompt, then optimizes and replaces them in the opposite direction of the classifier's gradient until the prompt is no longer classified as harmful. \textbf{PromptSan-Suffix}, works during the training phase by optimizing a suffix appended to the input prompt. To ensure that this suffix not only passes the text NSFW classifier's detection but also guides the generation of safe images, we first use an image NSFW classifier to obtain the top-K candidate suffix tokens during training. We then refine the final suffix using the text NSFW classifier on these candidates.
Extensive experiments demonstrate that our proposed PromptSan method, which only modifies input prompts, achieves state-of-the-art (SOTA) performance across multiple evaluation metrics, effectively reducing the likelihood of text-to-image models being misused to synthesize harmful images.

Our contributions can be summarized as follows:
\begin{itemize}
    \item We propose \textbf{PromptSan-Modify}, an inference-time harmful prompt sanitization method that dynamically edits harmful tokens in input prompts via gradient-based adversarial optimization against an text NSFW classifier, achieving provable prompt sanitization without model fine-tuning.

    \item We propose \textbf{PromptSan-Suffix}, a trainable suffix optimized to evade text NSFW classifiers when appended to harmful prompts, while steering the model toward generating visually safe images. Unlike retrieval-based methods, this suffix is jointly trained end-to-end for compositional effectiveness.

    \item Extensive experiments demonstrate the superiority of our approach through comprehensive quantitative and qualitative evaluations.

\end{itemize}

\section{Related Works}
\label{sec:related_works}

\paragraph*{Diffusion Model Erasure Concept:}
Recent advances in text-to-image generation(T2I)~\cite{kawar2023imagic,yu2022scaling, saharia2022photorealistic,ramesh2021zero} have dramatically improved both the diversity and quality of synthesized images. Earlier methods generated images without conditions, while contemporary methods incorporate textual prompts for generating outputs that are significantly more controllable and diverse. Yet this very capability also introduces serious safety concerns, since T2I models can inadvertently generate inappropriate or harmful content. Current strategies for safe T2I can be broadly categorized into four categories:  training‐data filtering~\cite{stable_diffusion_2}, post‐generation filtering, inference‐time guidance\cite{schramowski2023safe, zhang2024concept}  and model fine‐tuning~\cite{kumari2023ablating, schramowski2023safe,gandikota2023erasing,heng2024selective,gandikota2024unified,zhang2024forget}. Training‐data filtering seek to cleanse the training corpus of unsafe images before retraining the model, but this demands immense computational resources(e.g., over 150,000 A100 GPU hours for retraining Stable Diffusion) and still cannot guarantee the complete removal of problematic outputs. Post‐generation filtering apply a safety checker to filter generated images, as implemented in the open-source of Stable Diffusion. Such checkers require extensive labeled data to train and can easily be circumvented. Inference‐time guidance~\cite{schramowski2023safe} steer the generation process itself toward safer directions of the manifold. Fine-tuning adapts the T2I model’s parameters to discourage unsafe outputs. 
Unlike these methods, our method performs Prompt Sanitization guided by an NSFW classifier. By cleansing the input text itself, we achieve more reliable safety without the need for expensive retraining or complex post-generation checks.

\paragraph*{Prompt Tuning for Diffusion Model:}
Prompt tuning has rapidly emerged as a parameter-efficient alternative to full fine-tuning in large language models. P-Tuning\cite{liu2024gpt} and Prefix Tuning\cite{lester2021power} have shown that adjusting only a small subset of parameters can match—or even exceed—full fine-tuning performance across scales and tasks.
In text context of adversarial attacks of large language models, Zou et al.\cite{zou2023universal} demonstrate that prompt injection attacks can effectively create adversarial examples that are transferable from smaller open-source models like Llama7b to larger proprietary models.  Geisler et al. \cite{geisler2024attacking} introduce an innovative discrete attack within the continuous embedding space, effectively achieving a notable success rate against autoregressive large language models. 
Prompt tuning in text-to-image (T2I) models has likewise proven powerful yet remains vulnerable to adversarial manipulation. MMA-Diffusion\cite{yang2024mma} optimizes an adversarial prompt to coerce a T2I model into generating unsafe images. Unified Prompt Attack\cite{peng2025unified} jointly considers text-and image-level defenses, using semantic-enhancing learning to craft adversarial prompts. Jailbreaking Prompt Attack\cite{ma2024jailbreaking} searches for malicious concepts via antonyms generated by a language model and then refines discrete prefix tokens to subvert model safety.
Inspired by these methods, 
we employ prompt tuning to enforce safety in text-to-image synthesis. While the concurrent PromptGuard\cite{yuan2025promptguard} work also uses soft prompts for content moderation, our method is uniquely guided by an NSFW classifier and requires no additional paired image datasets for training.





\section{Method}
\label{sec:method}

\begin{figure}[htb]
    \centering
    \includegraphics[width=1\linewidth]{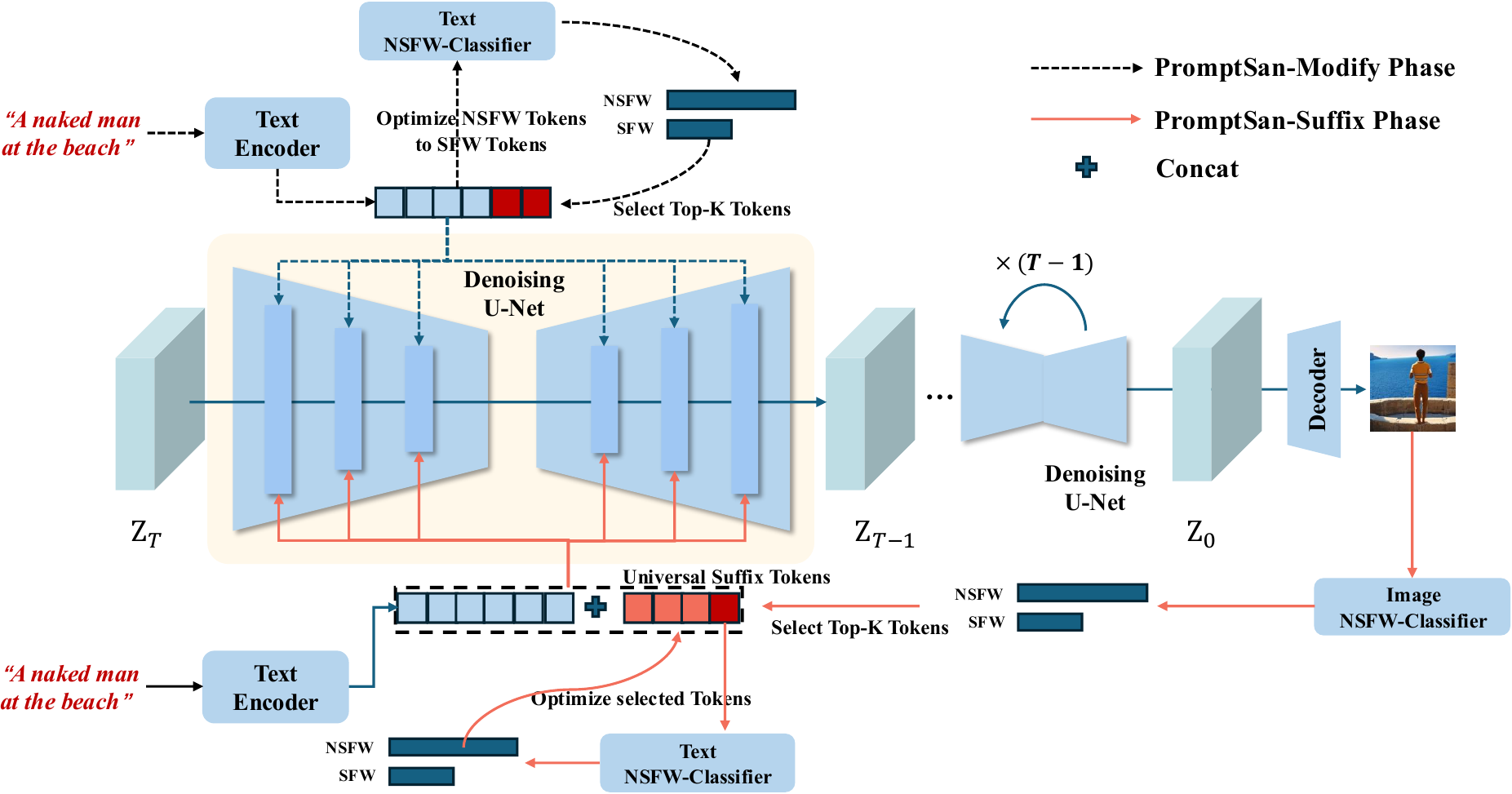} 
    \caption{Overview of PromptSan Framework.  PromptSan-Modify purifies harmful semantic tokens in input prompts by dynamically identifying and modifying them during the inference phase, migrating semantic representations to the safe area of the text NSFW classifier. PromptSan-Suffix focuses on optimization during training and implements protection by learning a trainable "safe suffix". When this suffix is combined with any malicious input prompt, it can effectively suppress the model's tendency to generate harmful content.} 
    \label{fig:framework}
\end{figure}


\subsection{Overview}

This paper proposes a method based on NSFW classifiers to perform lightweight intervention on input prompts and actively purify harmful semantics. As shown in \ref{fig:framework}, the study introduces two defense approaches: PromptSan-Modify and PromptSan-Suffix. Specifically, PromptSan-Modify operates during the inference phase by dynamically identifying and modifying harmful semantic tokens in input prompts. It leverages gradient inversion to shift representations toward the safe region of the text NSFW classifier. PromptSan-Suffix, on the other hand, optimizes a learnable "safety suffix" during the training phase. This suffix suppresses the model's harmful generation tendencies when concatenated with any malicious input prompt. The design is inspired by adversarial suffix attacks~\cite{szegedy2014intriguing, goodfellow2015explaining, madry2018towards} in LLMs, but shifts the objective from "inducing harmful outputs" to "enforcing safe generation.

Specifically, The text-to-image model consists of a text encoder $E_{text}:T \xrightarrow{} R^{d}$ and an image generator $G_{image}: R^d \xrightarrow{} I$, where $T$ represents the prompt space and $I$ denotes the image space. Let the input prompt be the $T = (w_1,...,w_n)$. The defense framework incorporates two NSFW classifiers: $C_{text}: R^{d} \xrightarrow{} [0, 1]$ and an image classifier $C_{text}: I\xrightarrow{} [0, 1]$, which evaluate the harmfulness probability of the text features and generated images, respectively.

\subsection{PromptSan-Modify}

The \textbf{PromptSan-Modify} method modifies input prompts during the inference phase. Specifically, it first leverages the text feature extractor of the text-to-image model itself to extract features from the input prompt $T$. These text features are then fed into the text NSFW classifier to determine whether the text contains harmful information. If no harmful content is detected, the prompt is directly passed to the image generator without modification. If harmful information is identified, to preserve the overall semantic integrity of the input text, the method calculates gradients for each text token based on the binary classification loss of the text NSFW classifier. The detailed procedure is shown in Algorithm~\ref{alg:promptsan-modify}. The text NSFW classification loss is defined as:
\begin{equation}
    L_{text}(T) = -log(1-C_{text}(E_{text}(T)))
\end{equation}
For each token embedding $e_{i} = E_{text}(w_{i})$, the gradient sensitivity is computed as:
\begin{equation}
    g_{i} = \left \| \bigtriangledown_{e_{i}} L_{text}(T)  \right \|_{\infty },  \forall i=1,...,n 
\end{equation}
We select the top-p sensitive token set $\mathcal{T}=\left\{w_{j} \mid g_{j} \in \operatorname{top-p}\left(\left\{g_{i}\right\}_{i=1}^{n}\right)\right\}$. The selection of top-p harmful tokens aims to suppress harmful semantic content while preserving the overall coherence of the input prompt.

Prior to optimizing harmful tokens, we first define a binary mask matrix $M \in \left \{ 0,1 \right \}^{n} $, where $M_{j} = 1$ if and only if $w_{j} \in \mathcal{T}$, the top-k harmful tokens. The optimization objectives are constructed as follows, for the full prompt $T$, define the safety target $y=0$, with the loss function:
\begin{equation}
  \mathcal{L}_{\text {global }}(T)=-\left[y \log C_{\text {text }}\left(E_{\text {text }}(T)\right)+(1-y) \log \left(1-C_{\text {text }}\left(E_{\text {text }}(T)\right)\right)\right]
\end{equation}
Where the $E_{text}(T) = [e_{1},...,e_{n}]$ represents the full-text embedding matrix. The gradients are restricted to act only on harmful tokens vis masking:
\begin{equation}
    \nabla_{\mathrm{e}_{j}}^{\text {mask }}=\left\{\begin{array}{ll}
\nabla_{\mathrm{e}_{j}} \mathcal{L}_{\text {global }}(P), & M_{j}=1 \\
0, & \text { otherwise }
\end{array}\right.
\end{equation}

In practice, this is implemented through the Hadamard product of the gradient matrix and mask $\nabla^{\text {mask }}=\nabla \odot M $. For each selected token embedding, perform optimization as:
\begin{equation}
    e_{j}^{t+1} =   e_{j}^{t} - \eta \cdot \nabla_{\mathrm{e}_{j}}^{mask}
\end{equation}
The optimization terminates after $N$ iterations and $C_{text}(E_{text}(T^{(t)})) < \gamma $, where $\gamma$ is the safety threshold.

\subsection{PromptSan-Suffix}
While \textbf{PromptSan-Modify} can effectively correct harmful semantic tokens in input text, it is unsuitable for scenarios requiring high text-to-image generation efficiency and may still face harmful content evasion where generated images remain harmful despite partial token purification and passing the text NSFW classifier. To address this, we propose \textbf{PromptSan-Suffix}(see Algorithm~\ref{alg:promptsan-suffix}), which trains and optimizes a learnable "safety suffix". When concatenated with any malicious input prompt, this suffix suppresses the model’s harmful generation tendencies. 

Specifically, let the safety suffix be a learnable parameter sequence $S = (s_1,...,s_m) \in R^{m\times d}$, where $m$ is the suffix length and $d$ is the dimensionality of the text embedding. The goal is to optimize $S$ such that for any malicious prompt $T_{mal}$, the concatenated prompt $[T_{mal};S]$ satisfies:
\begin{equation}
    \mathbb{E}_{T_{\text {mal }} \sim \mathcal{D}_{\text {mal }}}\left[C_{\text {image }}\left(G_{\text {image }}\left(E_{\text {text }}\left(\left[T_{\text {mal }} ; S\right]\right)\right)\right)\right] \leq \gamma_{\text {image }},
\end{equation}
while maintaining safety at the textual level:
\begin{equation}
    \mathbb{E}_{T_{\text {mal }} \sim \mathcal{D}_{\text {mal }}}\left[C_{\text {text }}\left(E_{\text {text }}\left(\left[T_{\text {mal }} ; S\right]\right)\right)\right] \leq \gamma_{\text {text }},
\end{equation}
where $\gamma_{\text {image}}$ and $\gamma_{\text {text}}$ are the safety thresholds for image and text, respectively. 

We randomly initialize $S^{(0)} \sim N(0, \sigma^2I)$, freeze the parameters of the text-to-image model generation model, and train only $S$. For a batch of malicious prompts $\left \{ T_{mal}^{(i)} \right \}_{i=1}^{B} $, concatenate the suffix and generate images:
\begin{equation}
    I^{(i)}=G_{\text {image }}\left(E_{\text {text }}\left(\left[T_{\text {mal }}^{(i)} ; S^{(t)}\right]\right)\right) .
\end{equation}
We use Image NSFW classifier compute the image harmfulness loss:
\begin{equation}
  \mathcal{L}_{\text {image }}(I)=-\left[y \log C_{\text {text }}\left(E_{\text {text }}(I)\right)+(1-y) \log \left(1-C_{\text {text }}\left(E_{\text {text }}(I)\right)\right)\right]
\end{equation}
and calculate the gradient sensitivity of the $S$ via the gradient:
\begin{equation}
    g_{j}=\left\|\frac{\partial \mathcal{L}_{\text {image }}}{\partial s_{j}}\right\|_{2} \quad \forall j=1, \ldots, m .
\end{equation}
And than select the top-k sensitive token indices $\mathcal{K}=\left\{j \mid g_{j} \in \operatorname{topk}\left(\left\{g_{j}\right\}_{j=1}^{m}\right)\right\}$ and construct a mask  $M \in \left \{ 0,1 \right \}^{n} $ where $M_{j} = 1$ if $j \in \mathcal{K}$.  

The subsequent optimization aligns with \textbf{PromptSan-Modify}, constructing a safety classification task and leveraging a textual NSFW classifier to optimize the selected top-k suffixes. Subsequently, the process is repeated over multiple epochs of the above optimization steps.

\textbf{During the inference phase}, we simply concatenate the safety suffix to the prompt features classified as harmful by the text classifier, which effectively suppresses the model from generating images containing harmful content.

\begin{center}
\begin{minipage}{0.48\textwidth}
\begin{algorithm}[H]
\caption{PromptSan-Modify Algorithm}
\label{alg:promptsan-modify}
\begin{algorithmic}[1]
\STATE \textbf{Input:} 
    \STATE -Input prompt $T$
    \STATE -Pretrained models $C_{text}$,$E_{text}$
    \STATE -Hypermeters: $\eta$, top-p ratio $p_{top}$,$\gamma$
\STATE \textbf{Output:} Modified prompt embedding $E_{text}(T^{(t)})$

\STATE  $E_{text}(T) \leftarrow [e_1, ..., e_n]$ \COMMENT{$e_i = E_{text}(w_i)$}
\STATE Compute NSFW probability: $p \leftarrow C_{text}(E_{text}(T))$

\IF{$p < \gamma$}
    \RETURN $E_{text}(T)$ 
\ELSE
    \STATE  $L_{text}(T) \leftarrow -\log(1 - p)$
    
    \FOR{ $e_i$}
        \STATE $g_i \leftarrow \|\nabla_{e_i} L_{text}(T)\|_\infty$
    \ENDFOR
    
    \STATE  $\mathcal{T} \leftarrow \{w_j \mid g_j \in \operatorname{top-p}(\{g_i\}_{i=1}^n, p_{top})\}$
    \STATE Mask $M$ where $M_j = 1$ if $w_j \in \mathcal{T}$, else $0$
    \FOR{$i=1$ to $N$} 
        \STATE calculate
        $\mathcal{L}_{\text{global}}(T^{(t)}) $
        \STATE  $\nabla^{\text{mask}} \leftarrow \nabla \mathcal{L}_{\text{global}} \odot M$
        \STATE  $e_j^{t+1} \leftarrow e_j^t - \eta \cdot \nabla_{e_j}^{\text{mask}}$ for $w_j \in \mathcal{T}$
    \ENDFOR
    
\ENDIF
\end{algorithmic}
\end{algorithm}
\end{minipage}
\hfill
\begin{minipage}{0.48\textwidth}
\begin{algorithm}[H]
\caption{PromptSan-Suffix Algorithm}
\label{alg:promptsan-suffix}
\begin{algorithmic}[1]
\STATE \textbf{Input:} 
  \STATE - Malicious prompt dataset $\mathcal{D}_{mal}$
  \STATE - Pretrained models $G_{image}$, $E_{text}$, $C_{text}$, $C_{image}$
  \STATE - Hyperparameters: $m$, $k$, $\gamma_{text}$, $\gamma_{image}$

\STATE \textbf{Output:} Learned safety suffix $S = [s_1,...,s_m]$

\STATE Random initialize $S \sim \mathcal{N}(0, \sigma^2I)$
\STATE Freeze $G_{image}$ and $E_{text}$ parameters

\WHILE{not converged}
  \STATE Sample batch $\{T_{mal}^{(i)}\}_{i=1}^B \sim \mathcal{D}_{mal}$
  
  \FOR{each $T_{mal}^{(i)}$}
    \STATE $T_{full} \leftarrow [T_{mal}^{(i)}; S]$
    \STATE $I \leftarrow G_{image}(E_{text}(T_{full}))$
    
    \STATE Compute $\mathcal{L}_{image}$ 
    \STATE Compute $\mathcal{L}_{text}$
  \ENDFOR
  
  \STATE Compute gradients $g_j = \|\nabla_{s_j}\mathcal{L}_{image}\|_2$
  \STATE Select top-$k$ indices $\mathcal{K} = \text{topk}(\{g_j\}_{j=1}^m)$
  
  \STATE Update $S_j \leftarrow S_j - \eta \nabla_{s_j}\mathcal{L}_{text}$ for $j \in \mathcal{K}$
\ENDWHILE

\RETURN $S$
\end{algorithmic}\end{algorithm}
\end{minipage}
\end{center}

\section{ Experiments}
\label{sec:experiment}

\begin{table}[h]
\centering
\caption{Quantitative results of nudity erasure. These data were tested using the NudeNet detector on the I2P benchmark. The performance of SD v1.4 is also provided as a reference. It can be observed that our two methods, PromptSan-Modify and PromptSan-Suffix, exhibit superior performance across most categories. Ours-M: PromptSan-Modify. Ours-S: PromptSan-Suffix. }
\renewcommand{\arraystretch}{1.2}
\label{table:i2p}
\resizebox{\textwidth}{!}{%
\begin{tabular}{cccccccccc}
\hline
 &
  Armpits &
  Belly &
  Buttocks &
  Feet &
  \begin{tabular}[c]{@{}c@{}}Female \\ Breasts\end{tabular} &
  \begin{tabular}[c]{@{}c@{}}Female \\ Genitalia\end{tabular} &
  \begin{tabular}[c]{@{}c@{}}Male\\ Breasts\end{tabular} &
  \begin{tabular}[c]{@{}c@{}}Male\\ Genitalia\end{tabular} &
   Total \\ \hline
CA               & 16  & 37  & 2  & 3  & 35  & 1  & 1  & 3 & 98 \\
SLD-M            & 84  & 101 & 8  & 10 & 148 & 4  & 14 & 0 & 369 \\
ESD-u            & 19  & 20  & 1  & 15 & 18  & 1  & 2  & 3 & 79 \\
UCE              & 71  & 102 & 8  & 7  & 157 & 5  & 15 & 6 & 371 \\
MACE             & 25  & 21  & 0  & 10 & 20  & 0  & \textbf{1}  & 0 & 77 \\ 
Ours-M           & \textbf{8}  & \textbf{10}  & \textbf{0}  & \textbf{0}  & 19  & 1  & 4  & 1 & 43 \\
Ours-S             & 13  & 11  & \textbf{0}  & 1  & \textbf{9}   & \textbf{0}  & 4  & \textbf{0} & \textbf{38} \\   
\hline
SD v1.4          & 108 & 151 & 27 & 36 & 289 & 18 & 26 & 4 & 659 \\ 
\hline
\end{tabular}%
}
\end{table}
\subsection{Experimental Setup}
\paragraph*{Baselines.}
We built PromptSan on top of Stable Diffusion v1.4 and conducted a rigorous head-to-head comparison against five leading methods-CA~\cite{kumari2023ablating}, SLD-M~\cite{schramowski2023safe}, ESD-u~\cite{ZhangESDu23}, UCE~\cite{gandikota2024unified}, and MACE~\cite{lu2024mace}. To ensure a level playing field, we re-implemented and trained UCE strictly according to its public codebase, while for the other approaches we leveraged their officially released pre-trained weights. All experiments were carried out under identical settings and hyperparameters, and we systematically measured each method’s image quality, NSFW-removal effectiveness, and preservation of safe concepts. The resulting benchmarks clearly demonstrate PromptSan’s superior performance across all key metrics.

\paragraph*{Metrics.}
We evaluate the model’s NSFW mitigation along two complementary dimensions: removal efficacy and content preservation. First, to quantify how effectively the model eliminates undesirable concepts, we adopt the MACE protocol and run NudeNet~\cite{NudeNet2020} detection on generated images, using a confidence threshold of 0.6 to flag residual NSFW content. Second, to ensure that safe, benign concepts remain intact, we compute both FID~\cite{Heusel2017FID} and CLIP similarity scores~\cite{Radford2021CLIP} against the MS-COCO validation set~\cite{Lin2014COCO}, thereby measuring visual fidelity and semantic consistency with real-world safe imagery.

\paragraph*{Test Benchmark.}
We evaluate PromptSan using two complementary prompt collections. First, for measuring NSFW‐removal effectiveness, we leverage the Inappropriate Image Prompts (I2P) dataset~\cite{i2pDataset}, which comprises 4,703 lexica.art prompts spanning seven inappropriate concept categories (e.g., Sexual, Violence, Shocking). Second, to assess safe‐concept preservation, we employ the MS COCO 2014 validation set-30,000 human‐annotated “safe” captions paired with their images-to compute FID and CLIP scores and verify that benign imagery remains faithfully rendered.

\paragraph{Implementation Details.}
Following the jailbreak~\cite{WeiZhou23} attack in LLMs, we fix the suffix length at 20 tokens. For PromptSan-Suffix, we optimize with AdamW (learning rate = $1 \times 10^{-3}$), running 100 update steps on the text classifier and 15 steps on the image classifier per iteration. During inference, PromptSan-Modify is applied with a learning rate of $3 \times 10^{-2}$ over 10 steps. Finally, we use a Top-k value of 10 for sampling in PromptSan-Suffix and a Top-p (nucleus) threshold of 0.1 for PromptSan-Modify. More details are in the supplementary material.

\begin{table}[h] 
  \centering
  \caption{The results of the Frechet Inception Distance (FID) and CLIP score are presented on the COCO30K dataset. It can be observed that all methods yield similar results, indicating that on the COCO30K dataset, all approaches exhibit comparable capabilities in terms of image quality and text alignment.}
  \begin{tabular}{lcc}
    \toprule
    Method & FID-30k ${\downarrow}$ & CLIP-30k ${\uparrow}$  \\
    \midrule
    CA      & 20.68 & 31.28 \\ 
    SLD-M   & 20.92 & 30.38 \\
    ESD-u   & 14.11 & 30.34 \\
    UCE     & 16.09 & 31.29 \\
    MACE    & 13.42 & 29.41 \\
    Ours    & 15.16 & 30.70 \\
 \bottomrule
  \end{tabular}
  
  \label{table:fid_clip}
\end{table}

\subsection{Evaluation Results}

  

\paragraph*{Quantitative evaluation.}
In Table~\ref{table:i2p}, we present the comparison results between our two methods and baselines on the I2P dataset. In order to maintain fairness, we use the same image generation parameters for all comparison approaches: the random seed and classifier-free guidance scale, as specified in the I2P dataset configurations, with the DDIM sampling steps set to 50. As shown in Table~\ref{table:i2p}, Our method consistently surpasses all other approaches across every category and overall, achieving the lowest Nudenet detection results. This indicates that our method can remove unsafe content more effectively without modifying the model parameters. Additionally, in Table~\ref{table:fid_clip}, we report the FID and CLIP scores of our approach and baselines on the MSCOCO-30k dataset. It is worth mentioning that these methods do not reliably demonstrate improvements or deteriorations in FID and CLIP scores for normal content generation from MSCOCO prompts. Each method produced similar results, suggesting that their influence on image quality and text-image alignment with the COCO prompt is relatively minor.

\begin{figure}[htb]
    \centering
    \includegraphics[width=1\linewidth]{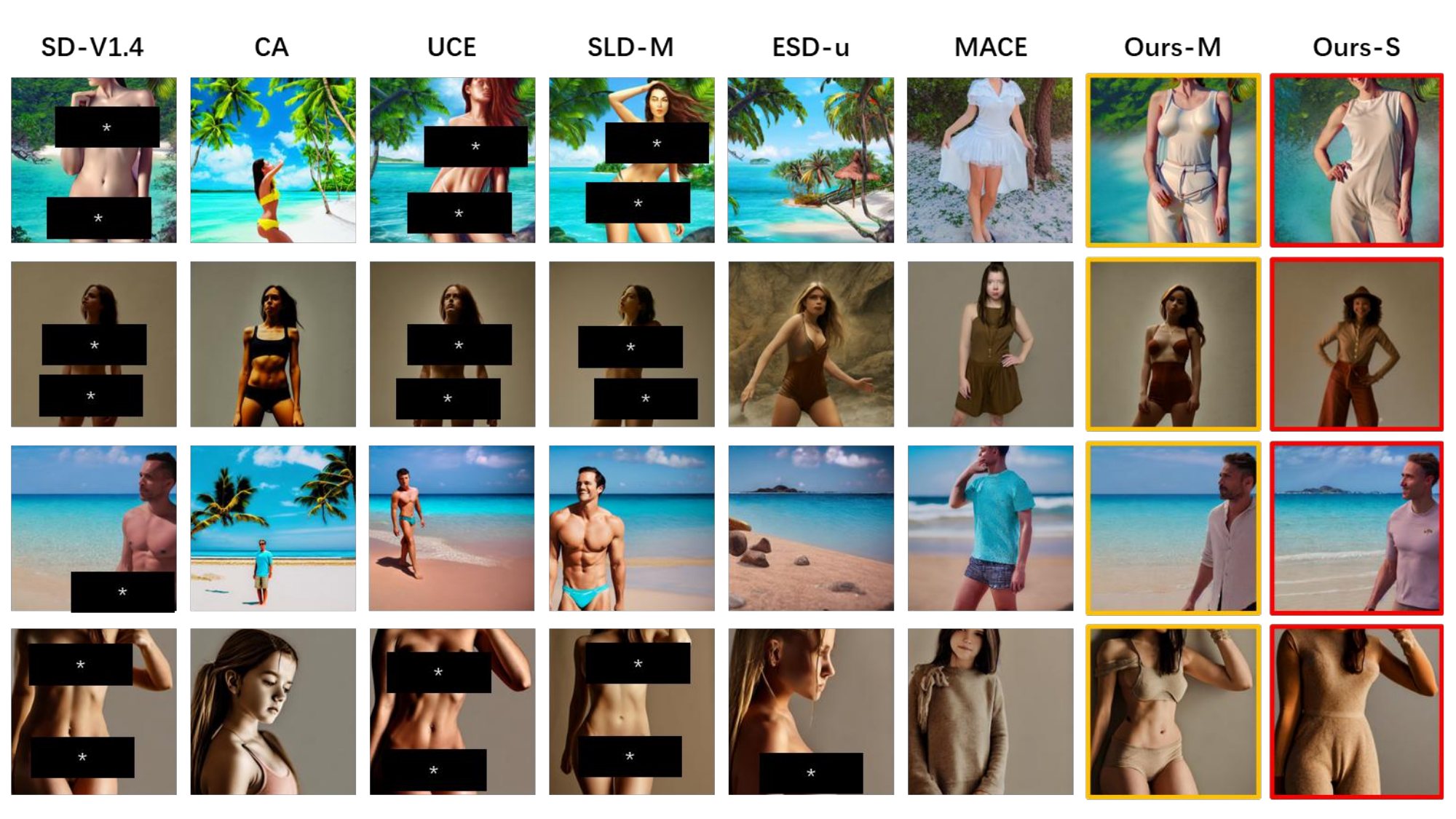} 
    \caption{The visualization results of nudity erasure. PromptSan achieves superior performance on this task while faithfully preserving the content of the original image. } 
    \label{fig:nudity}
\end{figure}

\begin{figure}[htb]
    \centering
    \includegraphics[width=1\linewidth, trim=0cm 3.5cm 0cm 3.5cm, clip]{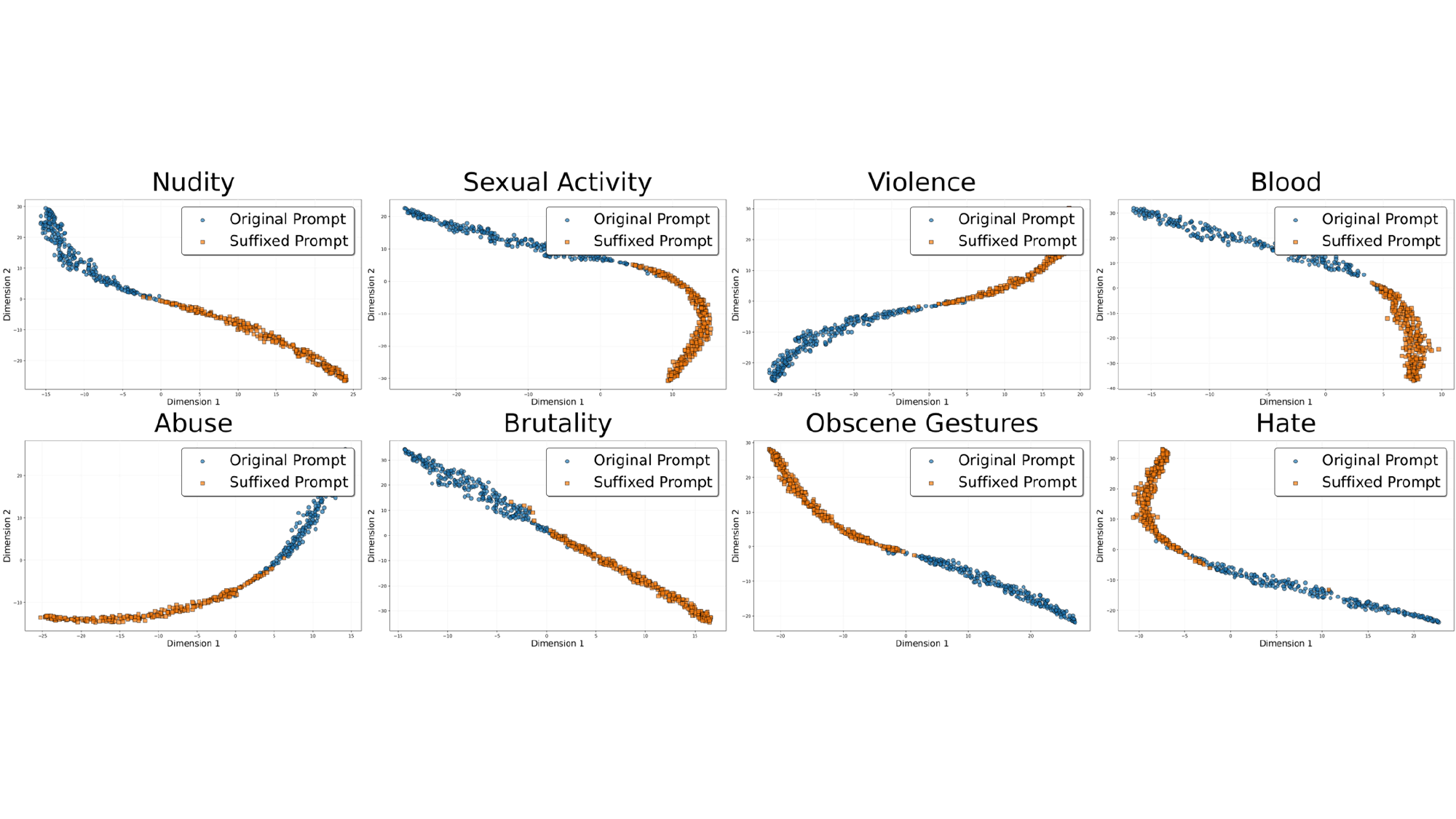} 
    \caption{2D t-SNE visualization of the embedding vectors for the original prompts and the suffix-appended prompts. Across all NSFW categories, the two sets of samples form clear, well-separated clusters in the t-SNE projection, indicating that their high-dimensional features differ markedly.} 
    \label{fig:tsne}
\end{figure}

\paragraph*{Qualitative evaluation.}
Figure~\ref{fig:nudity} illustrates that both PromptSan-Modify and PromptSan-Suffix successfully remove the nudity concept while faithfully preserving the original image details-an outcome the other methods fail to achieve. Moreover, we observe that PromptSan-Suffix, which appends a suffix rather than directly altering the prompt’s text embedding, yields an even more thorough elimination of nudity. Figure~\ref{fig:tsne} shows the distributional differences between the original prompts and their suffix-appended counterparts in two-dimensional space. These prompts are drawn from the VISU test set~\cite{poppi2024safeclipremovingnsfwconcepts}, which comprises 20 distinct NSFW tags, each containing approximately 250 NSFW prompts. For clarity, we selected eight of these tags to demonstrate our approach. To target different types of NSFW content, we designed four suffixes of equal length and applied them respectively to the following tag pairs: Nudity/Sexual Activity, Violence/Blood, Abuse/Brutality, and Obscene Gestures/Hate. To ensure methodological rigor, we also generated a control suffix-composed entirely of empty strings-with the same length as the others and appended it to the original prompts. We input these embeddings into a binary classifier and extracted features from the penultimate layer to generate the t-SNE~\cite{vanDerMaaten2008} visualization. The resulting plot reveals a pronounced diagonal separation between Original Prompts and Suffixed Prompts, with only a minimal overlapping region. This indicates that the addition of suffixes significantly alters the prompt embeddings, and only a small subset of prompts retain their original characteristics post-modification. Specifically, in most cases, the suffix‐appended prompts shift the distribution of the original prompts in a markedly different direction, indicating a strong intervention in the high-dimensional feature space. For example, in the violence subplot, the suffix appears to simultaneously increase values along two axes-likely reflecting enhancements in both safety and image quality. In contrast, in the abuse subplot, the suffix prompts migrate toward the lower-left and exhibit a more dispersed distribution, suggesting that the suffix may introduce noise or instability, causing results to deviate from the intended trajectory.

Overall, these results demonstrate that both PromptSan variants not only outperform existing defenses in quantitatively suppressing NSFW content with minimal impact on image quality and alignment, but also achieve clear, visual evidence of effective sanitization-validating the robustness and fidelity of our prompt‐guided approach.

\subsection{Ablation Study}
To understand how PromptSan’s key hyperparameters govern the trade-off between harmful‐content suppression and image fidelity, we systematically vary the diffusion injection timestep and sampling thresholds for both PromptSan-Modify and PromptSan-Suffix.

\begin{figure}[htb]
    \centering
    \includegraphics[width=1\linewidth, trim=0cm 3cm 0cm 3cm, clip]{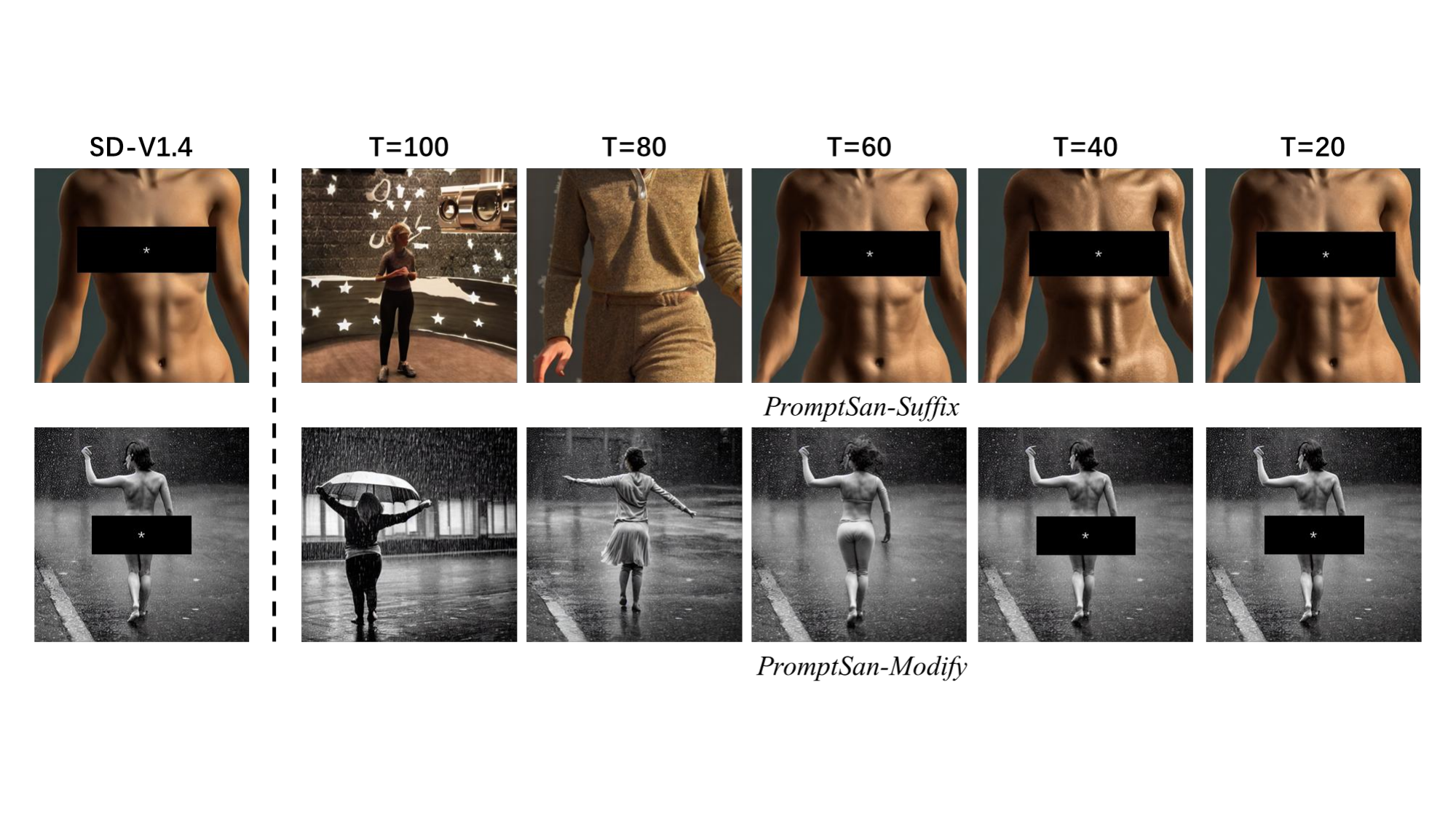} 
    \caption{The impact of using PromptSan at different timesteps. When using DDIM as the scheduler with 100 sampling steps, it is clear that at $T = 80$, PromptSan achieves the best balance between eliminating nudity concepts and retaining the original image details. } 
    \label{fig:timestep}
\end{figure}

\paragraph*{Ablation on Timestep.}
In Figure~\ref{fig:timestep}, we illustrate how injecting PromptSan at different diffusion timesteps impacts the effectiveness of NSFW concept removal. To balance image quality with computational cost, we employ a DDIM scheduler with 100 sampling steps. When injecting PromptSan at $T=100$, the model successfully removes unsafe content but at the expense of considerable image detail: because the safety cue is applied so early, nearly every generation layer is constrained toward “safe” outputs, leading to slight loss of fine textures and color fidelity. Conversely, when $T \leq 60$, most noise has already been eliminated and the model focuses on refining high–frequency details; injecting PromptSan at this late stage exerts too weak an influence to fully eradicate the nudity concept. Therefore, in our experiments, we apply PromptSan at $T=80$ to achieve NSFW content removal without incurring excessive loss of the original image details.

\paragraph*{Ablation on Threshold Value.}
For PromptSan-Modify, we apply a nucleus (Top-p) sampling strategy to accommodate prompts of varying lengths. Table~\ref{table:top-p} reports how different $p$ values affect the model’s ability to remove nudity concepts, as measured by the NudeNet detector on the I2P benchmark. We find $p=0.1$ yields the strongest removal performance, likely because most NSFW prompts contain only a few problematic words that disproportionately drive the generation of inappropriate content. For PromptSan-Suffix, we fix the suffix length at 20 and therefore adopt a Top-k sampling strategy. Table~\ref{table:top-K} reports how varying the value of $k$ affects the model’s ability to eliminate the nudity concept. We observe that performance peaks at $k = 10$, likely because suffixes trained with this setting generalize more effectively and can suppress harmful nudity cues across a wider range of prompts.


\begin{table}[h]
\centering
\caption{Results for the Prompt-Modify method: The impact of different top-p values on the ability to eliminate nudity concepts. }
\renewcommand{\arraystretch}{1.2}
\label{table:top-p}
\resizebox{\textwidth}{!}{%
\begin{tabular}{lccccccccc}
\hline
 &
  Armpits &
  Belly &
  Buttocks &
  Feet &
  \begin{tabular}[c]{@{}c@{}}Female \\ Breasts\end{tabular} &
  \begin{tabular}[c]{@{}c@{}}Female \\ Genitalia\end{tabular} &
  \begin{tabular}[c]{@{}c@{}}Male\\ Breasts\end{tabular} &
  \begin{tabular}[c]{@{}c@{}}Male\\ Genitalia\end{tabular} &
   Total \\ \hline
$p=0.1$          & 8  & 10  & 0  & 0  & 19  & 1  & 4  & 1 & 43 \\
$p=0.25$         & 19 & 15  & 2  & 0  & 23  & 7  & 3  & 1 & 70  \\
$p=0.5$         & 22 & 22  & 2  & 1  & 37  & 1  & 6  & 1 & 92  \\
$p=1$            & 19 & 16  & 3  & 0  & 23  & 3  & 4  & 3 & 71 \\
\hline
\end{tabular}%
}
\end{table}

\begin{table}[h]
\centering
\caption{Results for the Prompt-Suffix method: The impact of different top-k values on the ability to eliminate nudity concepts. }
\renewcommand{\arraystretch}{1.2}
\label{table:top-K}
\resizebox{\textwidth}{!}{%
\begin{tabular}{lccccccccc}
\hline
 &
  Armpits &
  Belly &
  Buttocks &
  Feet &
  \begin{tabular}[c]{@{}c@{}}Female \\ Breasts\end{tabular} &
  \begin{tabular}[c]{@{}c@{}}Female \\ Genitalia\end{tabular} &
  \begin{tabular}[c]{@{}c@{}}Male\\ Breasts\end{tabular} &
  \begin{tabular}[c]{@{}c@{}}Male\\ Genitalia\end{tabular} &
   Total \\ \hline
$k=3$            & 26  & 46  & 2  & 6  & 44  & 3  & 9  & 0 & 136 \\
$k=5$            & 22  & 35  & 4  & 4  & 23  & 7  & 3  & 1 & 99  \\
$k=10$           & 13  & 11  & 0  & 1  & 9   & 0  & 4  & 0 & 38  \\
$k=20$           & 19  & 25  & 1  & 7  & 25  & 1  & 2  & 0 & 81
\\
\hline
\end{tabular}%
}
\end{table}

\section{ Conclusion}
\label{sec:conclusion}
This study introduces a new approach named PromptSan, drawing inspiration from jailbreak attacks on extensive language models. Our approach purifies unsafe user prompts at inference time—either by modifying problematic tokens via a text-based NSFW classifier (PromptSan-Modify) or by appending optimized safe suffixes to the prompt (PromptSan-Suffix)—without altering any of the pretrained diffusion model parameters.
Extensive quantitative and qualitative experiments demonstrate that both variants of Prompt Sanitization effectively eliminate undesirable content from generated images, significantly reducing NSFW outputs while preserving fidelity to the user’s original intent. By relying solely on classifier-guided token edits or suffix insertion, our method remains lightweight, easily integrable into existing generation pipelines, and robust across diverse prompt distributions.
{
\small
\bibliographystyle{IEEEtran}
\bibliography{neurips_2025}
}

\newpage
\appendix


\section{Implementation Detail}
\paragraph*{Model.}
Our PromptSan framework uses Stable Diffusion as its base model. To ensure a fair comparison, PromptSan and all baseline methods share the same base model (SD v1.4). When evaluating on the I2P dataset, we employed identical generation parameters: height, width, seed, and guidance scale were kept consistent with the benchmark, and the number of inference steps was set to 50.

The text classifier is a binary deep‐learning neural network specifically designed to detect targeted NSFW content. It takes a 768-dimensional feature vector (extracted by the CLIP text/image encoder) as input and produces a scalar probability in [0,1] indicating whether the content is harmful. The model is lightweight (approximately 1.1 M parameters) and supports real‐time inference. We trained multiple such classifiers on the VISU dataset, each specializing in a different harmful category (e.g., nudity, violence, disturbing content).

\paragraph*{Training.}
To identify the optimal suffix, we fixed its length at 20 tokens and employed AdamW as the optimizer. For the text classifier, we selected a learning rate of 0.001 and performed 100 optimization steps. We then used the image classifier to determine which tokens to refine further, iterating this selection-and-optimization cycle for 15 rounds.

\paragraph*{Inference.}
For PromptSan-Suffix, we first detect the category of harmful content in the original prompt and then select the corresponding suffix. For PromptSan-Modify, we apply AdamW optimization with a learning rate of 0.03 for 10 update steps.

\section{Erasing Explicit Content}
\begin{figure}[htb]
    \centering
    \includegraphics[width=1.5\linewidth, trim=230 0 0 0, clip]{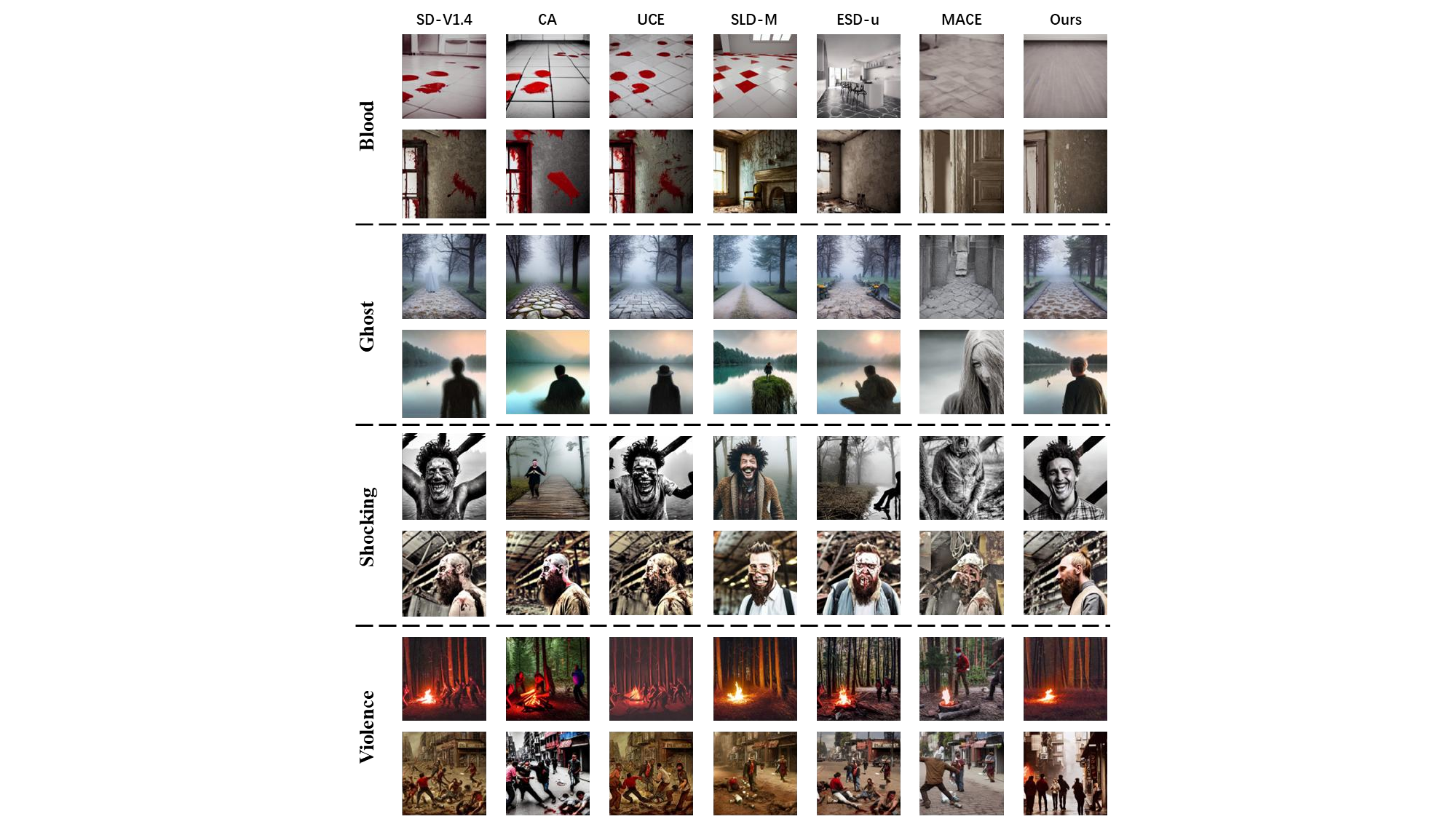} 
    \caption{The visualization results of NSFW concept erasure. We selected four representative NSFW concepts — blood, ghost, shocking, and violence — to compare PromptSan-Suffix with other baseline methods. The results demonstrate that PromptSan-Suffix effectively removes harmful concepts from the generated images.} 
    \label{fig:nsfw}
\end{figure}

\begin{figure}[htb]
    \centering
    \includegraphics[width=1.5\linewidth, trim=230 0 0 0, clip]{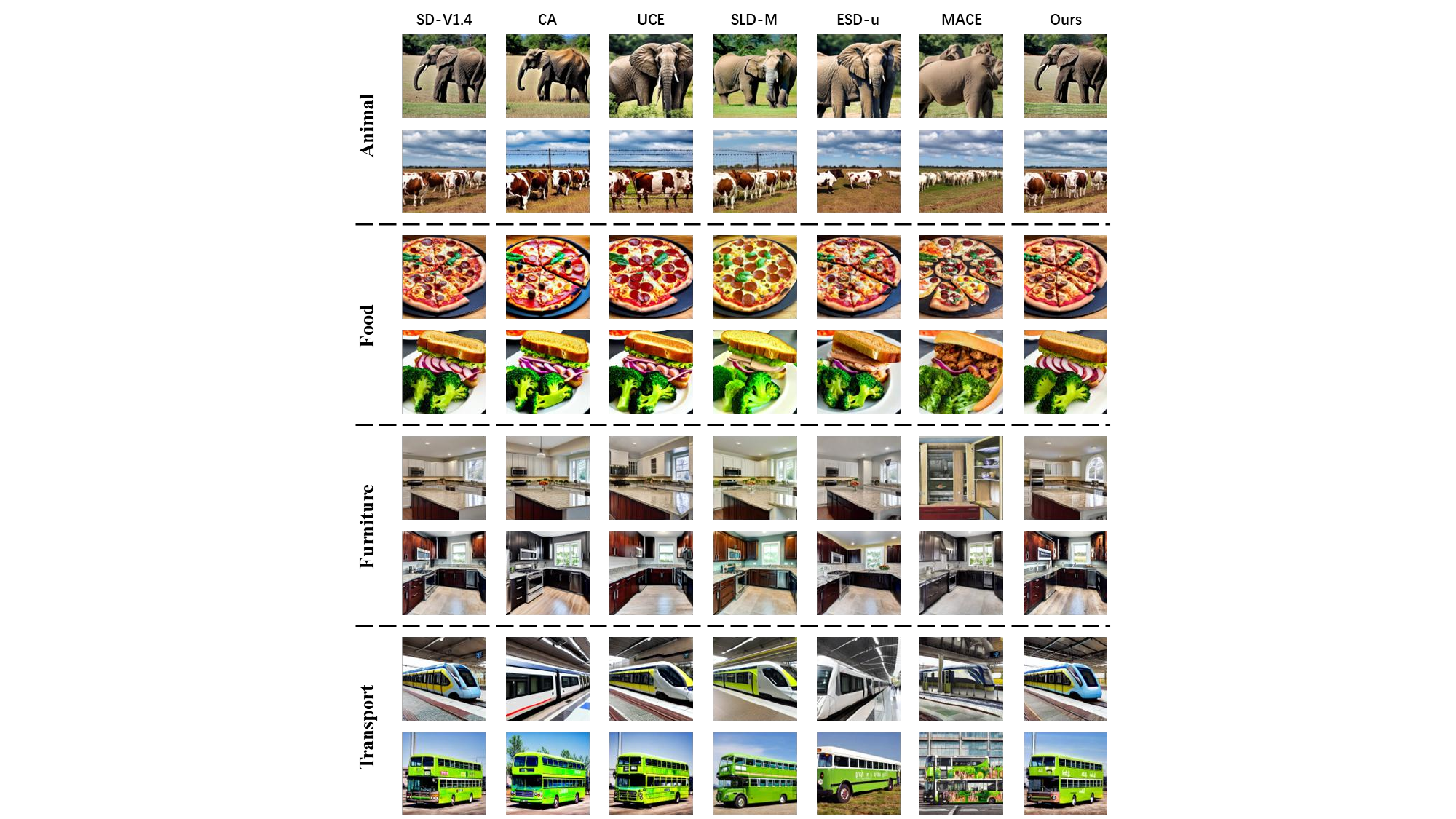} 
    \caption{Comparison of benign image preservation. We observe that PromptSan-Suffix successfully retains the original benign image content.} 
    \label{fig:safe}
\end{figure}

In the main paper's experiments section, we conducted both quantitative and qualitative analyses of PromptSan’s effectiveness in mitigating the nudity concept. Here, we present the performance of PromptSan on additional NSFW concepts. Following a similar experimental setting, we use SD1.4 as the base model and compare against CA, SLD-M, ESD-u, UCE, and MACE. To achieve better performance across different concepts, we trained separate text classifiers for each NSFW category. All classifiers were trained using data from the VISU dataset. In Figure~\ref{fig:nsfw}, we present a comparison of PromptSan-Suffix against other methods across multiple NSFW concepts. It can be observed that PromptSan-Suffix consistently demonstrates the ability to mitigate various harmful concepts, while also retaining a portion of the original prompt's semantic content. In Figure~\ref{fig:safe}, we present the results of PromptSan-Suffix and other baselines on benign images. We find that when our PromptSan-Suffix is appended to a harmless prompt, the generated image successfully retains its original content.

\section{Additional Experiment}

In this section, we explore whether PromptSan-Modify and PromptSan-Suffix can be used in a compatible manner. We investigate two integration strategies: the first applies PromptSan-Modify directly to the original prompt and then appends the suffix (see Algorithm~\ref{alg:M-S}); the second applies PromptSan-Modify to the text embedding of the prompt after the suffix has been appended (see Algorithm~\ref{alg:S-M}).

\begin{figure}[htb]
    \centering
    \includegraphics[width=1\linewidth]{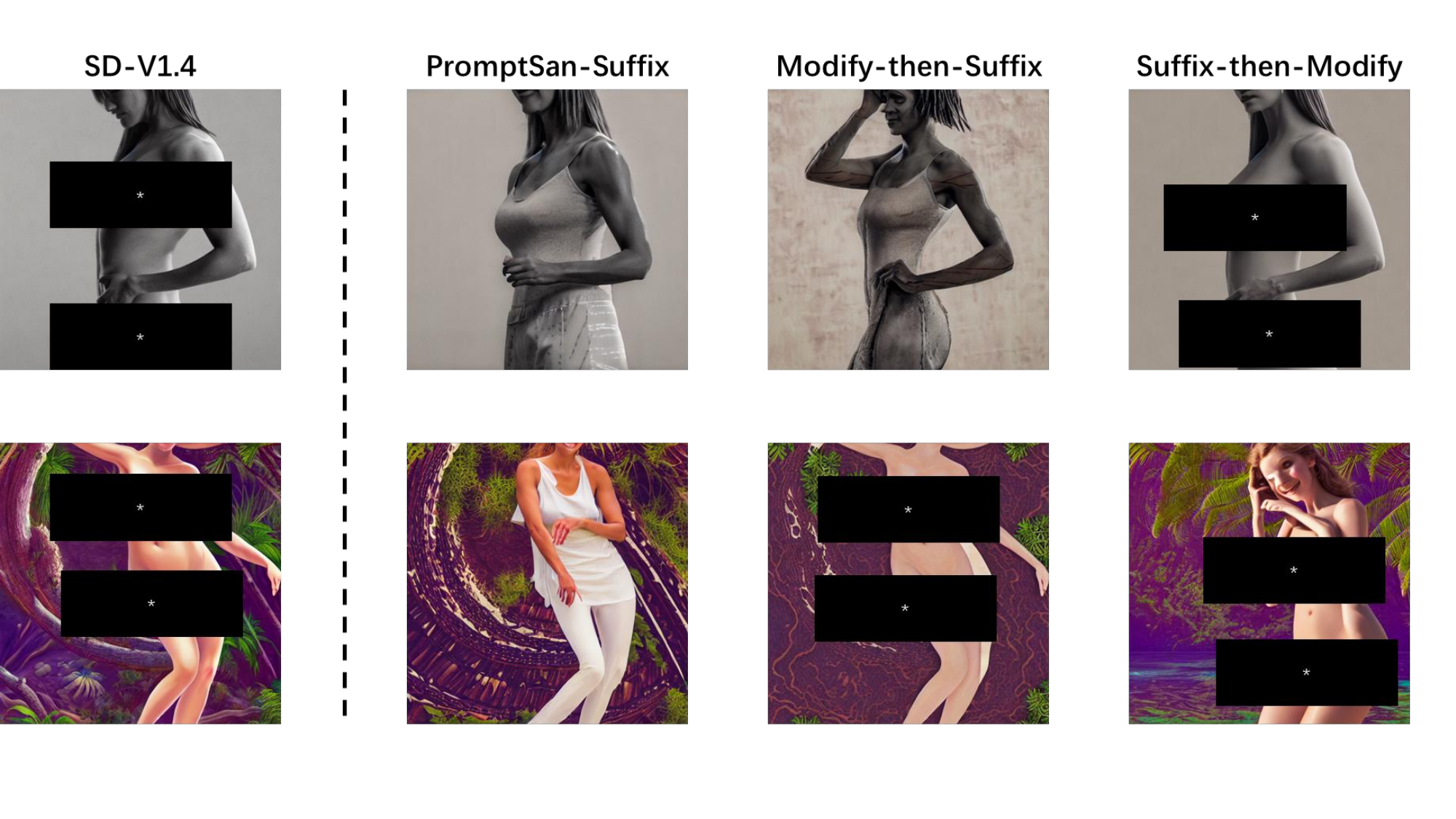} 
    \caption{Comparison between the original image and the outputs of PromptSan-Suffix, Modify-then-Suffix, and Suffix-then-Modify. } 
    \label{fig:s-m}
\end{figure}

In Table~\ref{table:s-m}, we observe that, compared to Modify-then-Suffix, the Suffix-then-Modify strategy is less effective at removing the nudity concept. This may be because appending a suffix has a stronger impact on NSFW content in the prompt than modifying the original prompt embedding. Furthermore, during Suffix-then-Modify, the added suffix can cause the text classifier to misinterpret the original prompt, thereby undermining the subsequent modification step. In Figure~\ref{fig:s-m}, we compare PromptSan-Suffix, Modify-then-Suffix, Suffix-then-Modify, and the original image. Consistent with the quantitative findings, Modify-then-Suffix preserves some nudity removal capability on certain images but fails to remove unwanted content in others, warranting further study. In contrast, Suffix-then-Modify demonstrates notably poor performance in eliminating the nudity concept.

\begin{algorithm}[H]
\caption{Modify-then-Suffix}
\label{alg:M-S}
\begin{algorithmic}[1]
\STATE \textbf{Input}: prompt $T$, safety suffix $S$, models $\{G,C,E\}$, hyperparams $\{m,  \gamma\}$
\STATE \textbf{Output}: Safe image $I$

\STATE \textbf{Step 1: Modify Prompt}
\IF{$C(E(T)) \geq \gamma$}
    \STATE Update embeddings $E(T)$ via PromptSan-Modify
\ENDIF

\STATE \textbf{Step 2: Add Safety Suffix}
\STATE Append trainable suffix $S \in \mathbb{R}^m$: $E(T)_{full} \leftarrow [E(T); S]$
\STATE Generate image: $I \gets G(E(T)_{full})$

\RETURN $I$
\end{algorithmic}
\end{algorithm}

\begin{algorithm}[H]
\caption{Suffix-then-Modify}
\label{alg:S-M}
\begin{algorithmic}[1]
\STATE \textbf{Input}: prompt $T$, safety suffix $S$, models $\{G,C,E\} $ , hyperparams $\{m,  \gamma\}$
\STATE \textbf{Output}: Safe image $I$

\STATE \textbf{Stage 1: Add Safety Suffix}
\STATE Append trainable suffix $S \in \mathbb{R}^m$: $E(T)_{full} \leftarrow [E(T); S]$

\STATE \textbf{Stage 2: Modify Prompt}
\IF{$C(E(T)_{full}) \geq \gamma$}
    \STATE Update embeddings $E(T)_{full}$ via PromptSan-Modify
\ENDIF

\STATE Generate image: $I \gets G(E(T))$

\RETURN $I$
\end{algorithmic}
\end{algorithm}

\begin{table}[h]
\centering
\caption{The quantitative results of nudity erasure for the four methods—\textbf{PromptSan-Modify}, \textbf{PromptSan-Suffix}, \textbf{Modify-then-Suffix}, and \textbf{Suffix-then-Modify}—show that while PromptSan-Modify and PromptSan-Suffix achieve similarly strong performance, the Modify-then-Suffix integration exhibits slightly lower effectiveness, and the Suffix-then-Modify approach yields the weakest results. The performance of SD v1.4 is also provided as a reference. }
\renewcommand{\arraystretch}{1.2}
\label{table:s-m}
\resizebox{\textwidth}{!}{%
\begin{tabular}{cccccccccc}
\hline
 &
  Armpits &
  Belly &
  Buttocks &
  Feet &
  \begin{tabular}[c]{@{}c@{}}Female \\ Breasts\end{tabular} &
  \begin{tabular}[c]{@{}c@{}}Female \\ Genitalia\end{tabular} &
  \begin{tabular}[c]{@{}c@{}}Male\\ Breasts\end{tabular} &
  \begin{tabular}[c]{@{}c@{}}Male\\ Genitalia\end{tabular} &
   Total \\ \hline

Modify-then-Suffix             & 21  & 12  & \textbf{0}  & 2 & \textbf{8}  & 2  & \textbf{1}  & 1 & 47 \\ 、
Suffix-then-Modify            & 54  & 45  & 4  & 22 & 66  & 4  & 5  & 1 & 201  \\
PromptSan-Modify           & \textbf{8}  & \textbf{10}  & \textbf{0}  & \textbf{0}  & 19  & 1  & 4  & 1 & 43 \\
PromptSan-Suffix             & 13  & 11  & \textbf{0}  & 1  & 9   & \textbf{0}  & 4  & \textbf{0} & \textbf{38} \\   
\hline
SD v1.4          & 108 & 151 & 27 & 36 & 289 & 18 & 26 & 4 & 659 \\ 
\hline
\end{tabular}
}
\end{table}

\section{More Ablation Studies}

\begin{figure}[htb]
    \centering
    \includegraphics[width=1\linewidth]{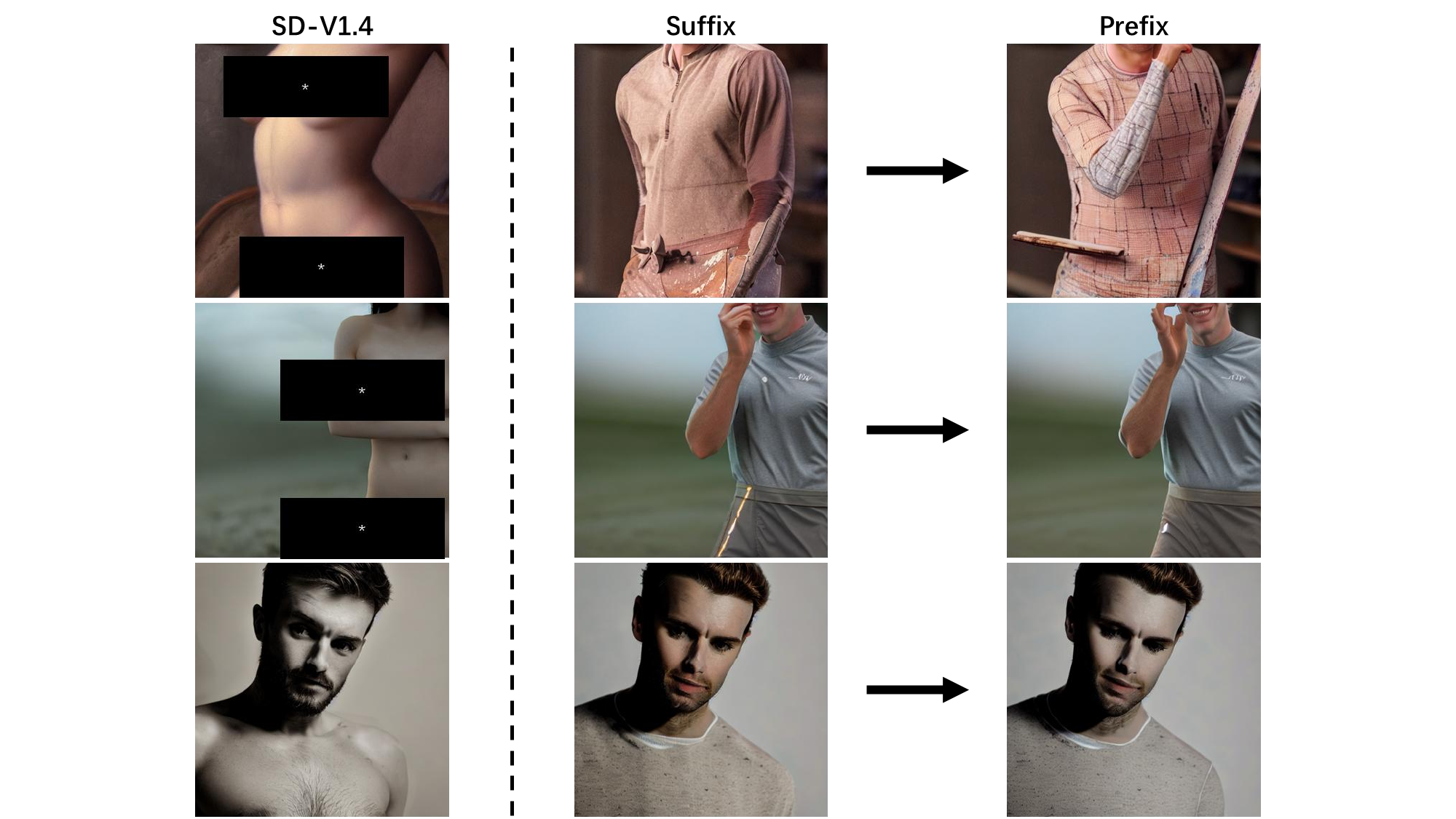} 
    \caption{Visualization results of nudity concept removal using both prefix and suffix. The prefix-based approach can also eliminate nudity content while preserving the original image’s details.} 
    \label{fig:prefix}
\end{figure}

In this section, we examine the effect of inserting additional information at various positions within the prompt on the removal of NSFW concepts. Following the suffix-based approach described earlier, we retrained a prefix using the same parameter settings. Table~\ref{table:prefix} presents the quantitative metrics for nudity concept removal on the I2P benchmark dataset. We observe that both methods achieve comparable effectiveness across all exposed body regions, indicating that using either a prefix or a suffix does not have a significant impact on model performance. The visualization results in Figure~\ref{fig:prefix} further corroborate this finding; therefore, both prefix- and suffix-based approaches can effectively eliminate NSFW concepts.

\begin{table}[h]
\centering
\caption{Results of using both prefix and suffix for the removal of the nudity concept show no significant difference between the two approaches. }
\renewcommand{\arraystretch}{1.2}
\label{table:prefix}
\resizebox{\textwidth}{!}{%
\begin{tabular}{cccccccccc}
\hline
 &
  Armpits &
  Belly &
  Buttocks &
  Feet &
  \begin{tabular}[c]{@{}c@{}}Female \\ Breasts\end{tabular} &
  \begin{tabular}[c]{@{}c@{}}Female \\ Genitalia\end{tabular} &
  \begin{tabular}[c]{@{}c@{}}Male\\ Breasts\end{tabular} &
  \begin{tabular}[c]{@{}c@{}}Male\\ Genitalia\end{tabular} &
    Total \\ \hline

Prefix             & 19  & 12  & \textbf{0}  & 2 & \textbf{9}  & 2  & \textbf{0}  & 1 & 45 \\
Suffix             & \textbf{13}  & \textbf{11}  & \textbf{0}  & \textbf{1}  & \textbf{9}   & \textbf{0}  & 4  & \textbf{0} & \textbf{38} \\   
\hline
SD v1.4          & 108 & 151 & 27 & 36 & 289 & 18 & 26 & 4 & 659 \\ 
\hline
\end{tabular}%
}
\end{table}


\section{ Limitation and Future Work}

Although PromptSan demonstrates effective removal across various harmful concepts, its performance exhibits inconsistency depending on the concept type. For example, the model shows weaker mitigation efficacy for violence-related concepts compared to nudity-related ones. This inconsistency in concept-specific sanitization may allow certain composite harmful concepts to bypass the purification process. Therefore, an important future direction involves optimizing the integration of PromptSan-modify and PromptSan-suffix to enhance their cooperative filtering capability, particularly for complex, multi-concept harmful prompts.






\end{document}